\documentclass[runningheads]{llncs}

\usepackage{eccv}

\usepackage{eccvabbrv}

\usepackage{graphicx}
\usepackage{booktabs}

\usepackage[accsupp]{axessibility}  

\usepackage{hyperref}

\usepackage{orcidlink}
\usepackage{pifont}
\usepackage{colortbl}
\usepackage{multirow}
\usepackage{bbding}

\usepackage{tabularx,booktabs}
\newcommand{\PreserveBackslash}[1]{\let\temp=\\#1\let\\=\temp}
\newcolumntype{C}[1]{>{\PreserveBackslash\centering}p{#1}}
\newcolumntype{R}[1]{>{\PreserveBackslash\raggedleft}p{#1}}
\newcolumntype{L}[1]{>{\PreserveBackslash\raggedright}p{#1}}

\begin{document}

\title{Knowledge-enhanced Visual-Language Pretraining for Computational Pathology} 

\titlerunning{KEP}

\author{Xiao Zhou$^1$,
\, Xiaoman Zhang$^{1,2}$,
\, Chaoyi Wu$^{1,2}$, \\[2pt]
\, Ya Zhang$^{1,2}$, 
\, Weidi Xie$^{1,2}$,
\, Yanfeng Wang$^{1,2}$}

\authorrunning{X. Zhou et al.}

\institute{Shanghai Artificial Intelligence Laboratory \and
Shanghai Jiao Tong University \\
\url{https://github.com/MAGIC-AI4Med/KEP} \\
\email{zhouxiao@pjlab.org.cn, \{xm99sjtu, wtzxxxwcy02, ya\_zhang, weidi, wangyanfeng\} @sjtu.edu.cn}}

\maketitle 

\begin{abstract}
  In this paper, we consider the problem of visual representation learning for computational pathology, by exploiting large-scale image-text pairs gathered from public resources, along with the domain-specific knowledge in pathology.
Specifically, we make the following contributions: 
(i) We curate a pathology knowledge tree that consists of 50,470 informative attributes for 4,718 diseases requiring pathology diagnosis from 32 human tissues. 
To our knowledge, this is the first comprehensive structured pathology knowledge base;
(ii) We develop a knowledge-enhanced visual-language pretraining approach, where we first project pathology-specific knowledge into latent embedding space via a language model, and use it to guide the visual representation learning;
(iii) We conduct thorough experiments to validate the effectiveness of our proposed components, demonstrating significant performance improvement on various downstream tasks, including cross-modal retrieval, zero-shot classification on pathology patches, and zero-shot tumor subtyping on whole slide images (WSIs). 
\end{abstract}

\section{Introduction}
\label{sec:intro}

Pathology diagnosis is currently the golden standard for examining various diseases in clinical applications, 
especially in the diagnosis of neoplasm~\cite{rorke1997pathologic}.
In the last decade, the prosperity of deep learning on computer vision has led to the rapid development of computational pathology, for example, approaches with supervised learning~\cite{shaban2020context,shao2021transmil,lin2023interventional,chan2023histopathology,huang2023conslide}, 
and weakly supervised learning~\cite{campanella2019clinical,zhou2020lirnet,lu2021data,chen2023rankmix,li2023task}.
Despite being promising, these approaches have been fundamentally limited by the scale of costly labels.
Alternatively, self-supervised pretraining on numerous unlabeled pathological images~\cite{wang2022transformer,chen2022fast,chen2022scaling,kang2023benchmarking} has attracted unprecedented attention, 
yet it still requires supervised fine-tuning for downstream deployments. 

\begin{figure}[t]
    \centering
    \includegraphics[scale = 0.6]{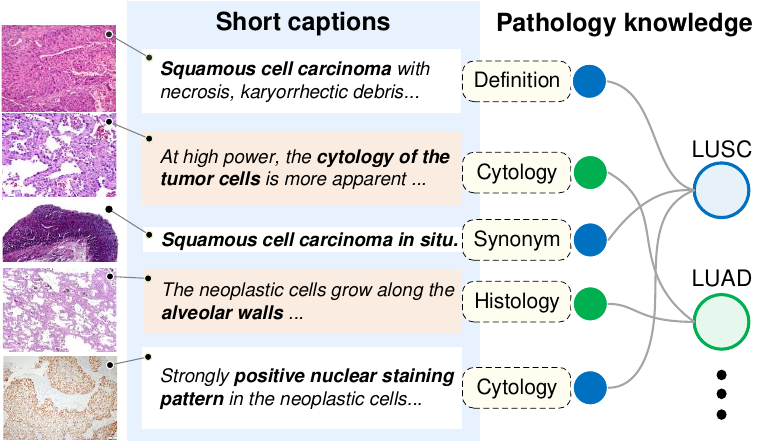}
    \caption{Knowledge-enhanced pathology image-text alignment.  The short caption of a pathology image crawled from public websites is typically unstructured and with varying granularities, which introduces noticeable ambiguities for image-text alignment. While the implicit structures and correlations between different image-caption pairs could be constructed by explicit disease attributes (dashed boxes), which can be well-aligned by a pathology knowledge tree. LUSC and LUAD suggest lung squamous cell carcinoma, and lung adenocarcinoma, respectively.
    } 
    \label{fig:0}
\end{figure}

In the recent literature~\cite{radford2021learning,jia2021scaling}, 
studies on the multimodal foundation model have demonstrated conspicuous improvement in downstream zero-shot tasks, 
by simply training to align visual and language embedding space on massive image-text pairs crawled from public websites. 
In contrast to computer vision~\cite{radford2021learning,jia2021scaling,hu2022st}, 
representation learning in pathology requires tremendous expertise and domain knowledge. An ideal training corpus would thus consist of well-structured medical reports from the hospital, however, it is often extremely difficult to acquire, due to privacy concerns. Alternatively, recent works~\cite{lu2023towards,lu2023visual,ikezogwo2024quilt,huang2023leveraging} propose to gather large-scale image-caption pairs from public resources~(PubMed~\cite{roberts2001pubmed} papers, Twitter, Youtube videos). Compared to discrete image labels, image captions from medical reports and academic articles can potentially provide more valuable medical information without manual annotation.

For existing works that adopt simple contrastive learning on pathology image and caption pairs, they suffer from the following challenges:
{\em First}, the short captions from these web-crawled image-text pairs are typically noisy, unstructured, and lack domain knowledge (Fig.~\ref{fig:0}), which can be sub-optimal for constructing high-quality pathology-specific visual representations; {\em Second}, the varying granularities of the free text from captions will inevitably introduce ambiguities when aligning with images, 
thus causing the model to be highly sensitive to the used text prompts during the inference stage~\cite{lu2023towards}. Thus, effectively harnessing the potential of such web-crawled data remains to be a challenge.

In this paper, to tackle the above challenges,
we anticipate that introducing pathology knowledge is of great significance to make up for the deficiency of short image captions.
To this end, we make the following contributions:
(i) We curate a pathology knowledge tree, 
\textbf{PathKT}, by collecting 50,470 informative pathological attributes of 4,718 diseases in 32 tissues from publicly available educational resources
and OncoTree\footnote{\label{oncotree}\url{https://oncotree.info/}}~\cite{kundra2021oncotree}.
(ii) We propose a novel knowledge encoder pretraining approach, that projects the attributes of each disease from PathKT into latent embedding space, where the attributes of the same disease, including disease synonyms, definitions, histology, and cytology features, share similar representations.
(iii) We develop a knowledge-enhanced pretraining (\textbf{KEP}) approach to align pathology visual-language representations, which freezes the knowledge encoder and continuously injects domain-specific knowledge into the image-text embedding space. 
To demonstrate the effectiveness of our proposed approach, we conduct thorough experiments on three downstream tasks, including retrieval on three pathological image-caption datasets, zero-shot patch classification on eight patch-level pathology image datasets, and zero-shot WSI tumor subtyping on three datasets from The Cancer Genome Atlas (TCGA)~\footnote{\url{https://portal.gdc.cancer.gov/}}. 
Quantitative experiments suggest that knowledge guidance can significantly enhance the performance across different tasks.

\section{Related work}

\noindent \textbf{Vision-Language Pretraining~(VLP).}
Current vision-language pretraining approaches are typically categorized into two groups. 
The first group, referred to as two-stream approaches~\cite{radford2021learning,xu2021videoclip}, involves using two separate encoders to extract features for visual and textual data, respectively. The second group, known as single-stream methods~\cite{kim2021vilt,chen2019uniter}, utilizes a cross-modal fusion encoder to enhance interactions between vision and text features.
In the field of medical VLP, existing methods mainly adopt the two-stream approach~\cite{zhang2020contrastive,huang2021gloria,chauhan2020joint,muller2022joint,lin2023pmc, zhang2023large,chen2022align}. Despite the valuable contributions of these methods, 
the reliance on data-driven representation learning restricts the utilization of systematic and structured medical knowledge, leading to less than professional diagnosis.

\vspace{3pt}\noindent \textbf{VLP in Computational Pathology.}
In the recent literature, Huang et al. proposed a pathology VLP model named PLIP~\cite{huang2023leveraging} and released OpenPath that contains 200K image-caption pairs from the social media of Twitter and other public resources. To extend the pathology data scale, Ikezogwo et al. curate Quilt1m~\cite{ikezogwo2024quilt} with over one million histopathology image-text pairs by capturing keyframes and speech from YouTube videos. In addition to evaluating the zero-shot classification on patch-level pathology images, Lu et al. propose MI-Zero~\cite{lu2023visual} and CONCH~\cite{lu2023towards} to extend the transfer ability of pathology VLP models on gigapixel whole slide images (WSIs). In contrast to these existing work, that finetunes a CLIP~\cite{radford2021learning} or CoCa~\cite{yu2022coca} on pathology image-caption pairs, 
we propose to first train a pathology-specific knowledge encoder, and use it to guide visual-language representation learning.

\vspace{3pt}\noindent \textbf{Medical Knowledge-enhanced Learning.}
In the medical community, leveraging external medical knowledge to enhance deep learning models has become an increasingly important topic~\cite{xie2021survey,wu2023medklip,zhang2023knowledge}. 
Generally, existing approaches can be categorized based on the ways of using medical knowledge: 
model-based approaches~\cite{huang2020dual,cui2020collaborative} adopts the prior knowledge of radiology or diagnosis summarized by doctors to design algorithms;
and input-based methods~\cite{chen2020generating,yu2021ernie,li2022cross,yang2022knowledge} directly exploit knowledge as the external input to guide representation training.
However, most of these works are focused on the analysis of chest X-rays.

\section{Methods}
\label{sec:methods}

Our primary goal is to leverage structured pathology knowledge to enhance visual-language representation learning.
To start with, we construct a \textbf{Path}ology \textbf{K}nowledge \textbf{T}ree, termed as \textbf{PathKT}, that consists of 50,470 informative attributes of 4718 diseases from 32 human tissues (Sec.~\ref{sec3.1.1}).
We then train a knowledge encoder that projects the structured pathology knowledge into an embedding space~(Sec.~\ref{sec3.1.2}).
We further employ the knowledge encoder to guide visual-language pretraining for computational pathology, termed as KEP (\textbf{K}nowledge-\textbf{E}nhanced \textbf{P}re-training, Sec.~\ref{sec3.2}).

\label{sec3.1}
\subsection{PathKT Construction}
\label{sec3.1.1}

Here, we detail the procedure for building up a pathology knowledge tree with various online sources, which will be further used for knowledge encoding. 

\begin{figure}[t]
    \centering
    \includegraphics[scale = 0.4]{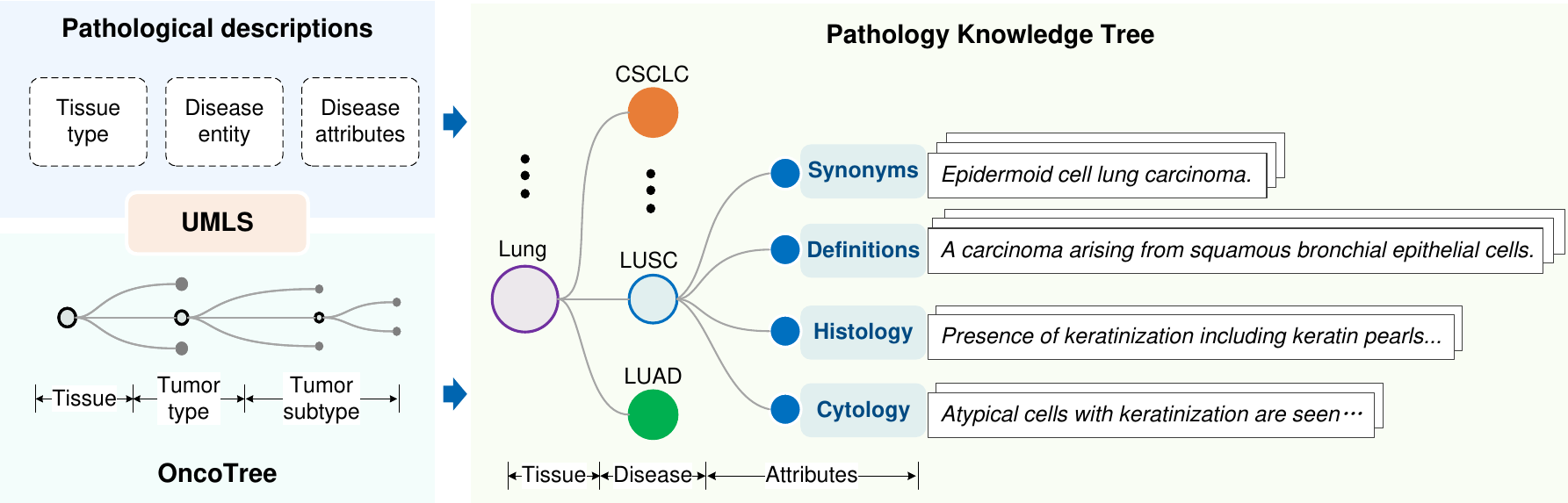}
    \caption{The construction of pathology knowledge tree. OncoTree is adopted as the base architecture to construct the PathKT. The tissue types, disease entities, and attributes are first extracted from web-crawled pathological descriptions, where cancers are then matched to OncoTree based on their tissue types and tumor types/subtypes using UMLS CUIs.
    Moreover, non-tumor diseases are added to the knowledge tree according to their tissue types.
    Finally, the pathology knowledge tree integrates 4718 diseases from 32 tissues, with each disease containing various synonyms, definitions, and histological and cytological features. CSCLC in this figure suggests combined small cell lung cancer.}
    \label{fig:1}
\end{figure}

\vspace{3pt}\noindent \textbf{Knowledge Source.}
We collect pathology-specific knowledge from publicly available educational resources, such as text books, professional websites, and structured databases (OncoTree~\cite{kundra2021oncotree}).
Specifically, we extract pathological descriptions of 884 tumor subtypes from OncoTree and 4360 diseases from text books and professional websites, including all domain knowledge required for diagnosis in clinical scenarios, {\em i.e.,} disease name/ synonyms, definitions, histological features, and cytological features.

\vspace{3pt}\noindent \textbf{Knowledge Tree Construction.}
We structure these knowledge sources into a knowledge tree by expanding the OncoTree, as shown in Fig.~\ref{fig:1}.
OncoTree is a tree-structure cancer classification system~\cite{kundra2021oncotree}, which consists of 884 tumor subtypes from 32 tissues, with each type of tumor linked to a Concept Unique Identifier (CUI) from Unified Medical Language System (UMLS)~\cite{bodenreider2004unified}. 
Specifically, we first extract tissue types, disease entities, and disease attributes from pathological descriptions of 4360 diseases, in which 168 cancers are found overlapped with OncoTree by using SciSpacy package~\cite{neumann-etal-2019-scispacy} to link the UMLS CUIs.
The rest 4192 diseases are then added to the knowledge tree according to their tissue types. 
After deduplication and noise reduction (358 diseases without any informative descriptions are deleted), 
all diseases are organized into the corresponding tissues.
The final histopathology knowledge tree contains 4718 diseases from 32 tissues.
The attributes of each disease node are constructed by a varying number of synonyms, definitions, and histological and cytological descriptions. The statistics of PathKT is shown in Table S1 and Fig. S1 in Supplementary Materials. In the final PathKT, for instance, the tissue node of the lung connects all lung diseases, including combined small cell lung cancer (CSCLC), lung squamous cell carcinoma (LUSC), and lung adenocarcinoma (LUAD), etc. The LUSC node connects four kinds of attributes, shown in Fig.~\ref{fig:1}.

\subsection{Pathology Knowledge Encoding}

\label{sec3.1.2}
In this section, we describe details for projecting tree-structure pathological knowledge into a latent embedding space, by training a knowledge encoder. 
Specifically, we align different disease entities with their corresponding free-text attributes via metric learning, such that,
the synonyms, definitions, and corresponding histological/cytological features are close in the embedding space.  
As a consequence, the model can link the pathology images to their implicit disease labels during the visual-language pertaining, since the free-form text descriptions in image-text pairs may contain disease attributes, such as pathological features and disease definitions, 
which have already been aligned with disease names/synonyms. 
Therefore, the alignment of pathology knowledge can help the model link to the disease entities for better performance during diagnosis.

\vspace{3pt} \noindent \textbf{Problem Setting.} 
Given a set of disease entities with their corresponding attributes, 
$\mathcal{D} = \{(d_1, \mathbf{a}_1), \dots, (d_n, \mathbf{a}_n)\}$, 
where $d_i$ denotes the $i$-th disease entity, 
and $\mathbf{a}_i = \{a_i^1, \dots, a_i^k\}$ refer to the associated $k$ attributes, both disease and attributes are represented in the format of natural language. Note that, for different disease entities, $k$ also varies,
our goal here is to train a model that satisfies:
\begin{align}
\small
\text{sim}(\Phi_{\text{k}}(a_i^p), \Phi_{\text{k}}(a_i^q)) \gg 
\text{sim}(\Phi_{\text{k}}(a_i^p), \Phi_{\text{k}}(a_j^t)), \text{ } i \neq j,
\label{eq:0}
\end{align}
where $\Phi_\text{k}(\cdot)$ denotes the knowledge encoder, 
$\text{sim}(\cdot, \cdot)$ refers to the similarity,
$a_i^p, a_i^q$ and $a_j^t$ refer to the randomly sampled attributes from the $i,j$-th disease entity. 
Intuitively, the knowledge encoder enables the attributes of the same disease to be pulled together, while attributes from different diseases are pushed apart.

\begin{figure}[t]
    \centering
    \includegraphics[scale = 0.54]{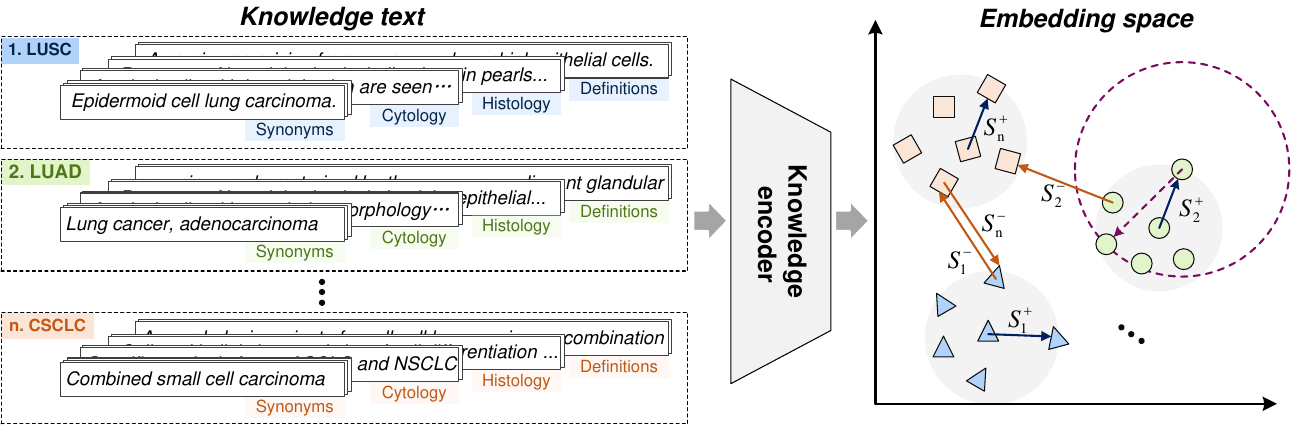}
    \caption{Knowledge encoder pretraining based on metric learning. $n$ disease entities and each with $k$ attributes, including disease synonyms, definitions, cytology and pathology features, construct a mini-batch (left part of the figure), which are fed to a knowledge encoder for pretraining. In the embedding space (right part of the figure), the markers in different shapes represent the embeddings of attributes of different diseases. $S^+_i$ suggests the max-min positive attribute similarity within the $i$-th disease, while $S^-_i$ denotes the maximal attribute similarity between the $i$-th disease and other diseases. The goal of metric learning is to increase $S^+_i$ and meanwhile decreasing $S^-_i$. The purple-dashed arrow and circle denote the minimal positive attribute similarity in the second class and the hypersphere it spans.}
    \label{fig:3}
\end{figure}

\vspace{3pt} \noindent \textbf{Training.} 
In order to achieve the objective defined in Eq.~\ref{eq:0}, metric learning is exploited to construct an embedding space where the representations of intra/inter-class instances are clustered/separated.
In specific, given a mini-batch that contains $n$ random diseases and each with $k$ attributes, we denote the normalized embedding for the $p$-th attribute of $i$-th disease as: $\mathbf{z}_{p}^i=\Phi_{\text{k}}(a_i^p)/\lVert \Phi_{\text{k}}(a_i^p)\rVert$.
We adopt the recently proposed AdaSP loss~\cite{zhou2023adaptive}, which finds out a max-min positive similarity (Fig.~\ref{fig:3}) and then shapes a loss with the maximal negative similarity:
\begin{equation}
\small
    \mathcal{L}_{\text{metic}}=\frac{1}{n}\sum_{i=1}^n{\log \left( 1+e^{(S_{i}^{-}-S_{i}^{+})/\tau} \right)},
    \label{eq:2}
\end{equation}
where $\tau$ is a temperature parameter. $S_{i}^{+}$ and $S_{i}^{-}$ denote the max-min positive and the maximal negative similarity, which can be computed by the soft version:
\begin{equation}\label{eq3}
\small
    S_i^{+} = \max_{p}{\min_{q}{\left< \mathbf{z}_p^i, \mathbf{z}_q^i \right> }} \approx \tau \log \left( \sum_{p=1}^k{\frac{1}{\sum_{q=1}^k{e^{-\frac{\left<\mathbf{z}_{p}^{i},\mathbf{z}_{q}^{i}\right>}{\tau}}}}} \right)
\end{equation}
\begin{equation}\label{eq3_add}
\small
    S_i^{-} = \max_{j,p,q}{\left< \mathbf{z}_p^i, \mathbf{z}_q^j \right> }\approx \tau \log \left( \sum_{p=1}^k{\sum_{j=1,j\ne i}^n{\sum_{q=1}^k{e^{\frac{\left<\mathbf{z}_{p}^{i},\mathbf{z}_{q}^{j}\right>}{\tau}}}}} \right) 
\end{equation}
where $\left< \cdot \right>$ represents the cosine similarity, the details about the soft version can be found in Supplementary Materials.

\vspace{3pt} \noindent \textbf{Discussion.} 
The conventional triplet loss~\cite{hermans2017defense} 
with batch hard mining strategy is widely adopted in metric learning, 
while it is not suitable in our case due to the following two reasons:
{\em First,} in the pathology knowledge, each disease entity is associated with at most four types of attributes, the text descriptions of different attribute types might reveal significant divergence, 
which causes a large intra-class variation, marked by the purple dashed circle in Fig.~\ref{fig:3}, during training. 
{\em Second,} the fine-grained diseases might share high similarities, 
such as tumor subtypes: breast invasive ductal carcinoma and breast invasive lobular carcinoma, which cause low inter-class variations during training. When a mini-batch meets these two conditions, traditional triplet loss that enforces the instance similarities could typically produce the bad local minima of optimization~\cite{xuan2020hard}, which, therefore, undermines the metric learning.

\subsection{Pathology Knowledge Enhanced Pretraining}
\label{sec3.2}
In this section, we present a simple yet effective pretraining approach, termed KEP, that leverages the established knowledge encoder to guide visual-language pretraining for computational pathology.

\begin{figure}
    \centering
    \includegraphics[scale = 0.36]{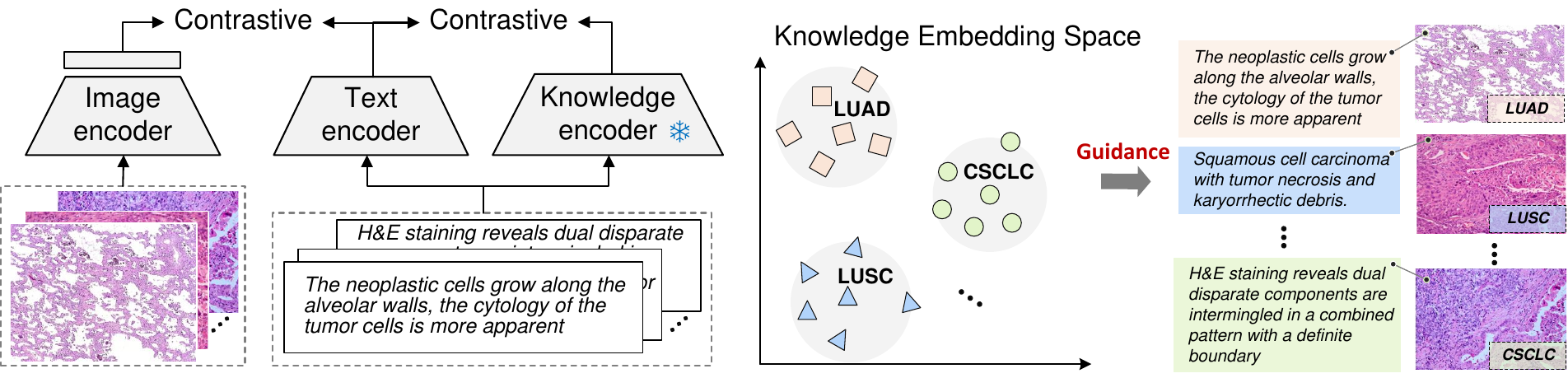}
    \caption{Model architecture (left graph). A projection head is added on the top of the visual encoder to bridge the gap between the image and the text encoder. The knowledge encoder is frozen across the whole training stage to distill pathology knowledge to the learnable text encoder. As a result, the pathology images can be aligned with their implicit disease labels ( marked by dashed boxes in the right graph) during visual-language pretraining, since the captions contain disease attributes that have been already aligned with disease names/synonyms in the knowledge embedding space.}
    \label{fig:4}
\end{figure}

\vspace{5pt} \noindent \textbf{Visual-Language Pretraining.}
Given paired image and captions, denoted as $\mathcal{F} = \{(x_1, c_1), \dots, (x_n, c_n)\}$, our goal is to construct a visual-language embedding space from paired image-text data, that satisfies:
\begin{align}
\small
\text{sim}(\Phi_{\text{v}}(x_i), \Phi_{\text{t}}(c_i)) \gg 
\text{sim}(\Phi_{\text{v}}(x_i), \Phi_{\text{t}}(c_j)), \text{ } i \neq j,
\label{eq:3}
\end{align}
where $\Phi_\text{v}(\cdot)$ and $\Phi_\text{t}(\cdot)$ denote the visual and the textual encoder, respectively. In this work, ViT-B-32/16 is adopted as the backbone for the visual encoder and initialized from the visual weights of CLIP~\cite{radford2021learning} / BiomedCLIP~\cite{zhang2023large}. 
A projection head is added on top of the visual encoder to bridge the gap caused by the non-pathology initialization, shown in Fig.~\ref{fig:4}.
The text encoder is initialized by the pretrained knowledge encoder, as described in Sec.~\ref{sec3.1.2}. 
In order to learn effective visual-language representation,
we optimise an infoNCE contrastive objective:
\begin{equation}
\small
   \mathcal{L}_{\text{vt}} =-(\log \frac{e^{\mathbf{v}_i^T\mathbf{t}_i/\tau}}{\sum_{j=1}^{n}e^{\mathbf{v}_i^T\mathbf{t}_j/\tau}} 
   + \log \frac{e^{\mathbf{t}_i^T\mathbf{v}_i/\tau}}{\sum_{j=1}^{n}e^{\mathbf{t}_i^T\mathbf{v}_j/\tau}}),
   \label{eq:4}
\end{equation}
where $\mathbf{v}_i = \Phi_{\text{v}}(x_i)$, and
$\mathbf{t}_i = \Phi_{\text{t}}(c_i)$, 
refer to the normalized embedding vectors 
from visual and text encoders, respectively. 
$\tau$ denotes a temperature parameter.

\vspace{5pt} \noindent \textbf{Knowledge Distillation.} 
To keep the alignment between images and free-form captions inside the knowledge space and thus the images can be linked to their implicit disease entities (right part in Fig.~\ref{fig:4}), we adopt an additional frozen branch to continuously distill pathology knowledge to the text encoder. 
Specifically, we use the text-knowledge embedding pairs to construct a contrastive loss item $\mathcal{L}_{\text{tk}}$, 
which can be computed by Eq.~\ref{eq:4} with the normalized visual embedding vectors replaced by the knowledge embedding vectors:
\begin{equation}
\small
   \mathcal{L}_{\text{tk}} =-(\log \frac{e^{\mathbf{k}_i^T\mathbf{t}_i/\tau}}{\sum_{j=1}^{n}e^{\mathbf{k}_i^T\mathbf{t}_j/\tau}} 
   + \log \frac{e^{\mathbf{t}_i^T\mathbf{k}_i/\tau}}{\sum_{j=1}^{n}e^{\mathbf{t}_i^T\mathbf{k}_j/\tau}}),
   \label{eq:5}
\end{equation}
where $\mathbf{k}_i = \Phi_{\text{k}}(c_i)$, refer to the normalized embedding vectors from the knowledge encoder.
The overall training loss can thus be computed as: 
\begin{equation}
\small
    \mathcal{L} = \mathcal{L}_{\text{vt}} + \alpha \mathcal{L}_{\text{tk}},
    \label{eq:6}
\end{equation}
where $\alpha$ denotes a weight parameter. It is worth emphasizing that the key contribution in KEP is to initialize the text encoder with the pre-obtained pathology knowledge encoder and adopt it to continuously distill pathology knowledge to the text encoder, while for the other parts, we keep the same as PLIP~\cite{huang2023leveraging}.

\section{Experiments}
In this section, we first introduce the datasets used for training and evaluation in this paper, followed by the evaluation metrics and implementation details.

\subsection{Training Datasets} 
\noindent \textbf{Dataset for Knowledge Encoding.} 
We carry out pathology knowledge encoding with PathKT, 
where 4,718 disease nodes with a total number of 50,470 attributes are extracted, shown in Table S1 in Supplementary Materials.

\vspace{3pt}\noindent \textbf{Datasets for KEP Pretraining.}
Our dataset is composed of two parts.
{\em First}, we collect the \textbf{OpenPath} data provided by PLIP~\cite{huang2023leveraging} and obtain 138,874 image-text pairs after denoising and pre-processing. 
{\em Second}, we collect the \textbf{Quilt1M}~\cite{ikezogwo2024quilt} dataset, which gathers pathology image-text pairs from four public sources: 
PubMed articles, LAION~\cite{schuhmann2022laion}, 
OpenPath~\cite{huang2023leveraging}, and Youtube videos.
Since one of the downstream evaluation datasets Arch-PubMed~\cite{gamper2021multiple} contains pathology images from PubMed articles, we remove this data source from Quilt1M to avoid data leaking.
Considering that many images in Quilt1M are matched with multiple captions, we concatenate the captions related to the same image and finally obtain 576,608 image-text pairs for the Quilt1M dataset.

\subsection{Downstream Tasks} 
We evaluate the pretrained models on three tasks, namely, retrieval, zero-shot patch classification, and zero-shot WSI tumor subtyping. The details for all evaluation datasets are exhibited in Table S2 in Supplementary Materials.

\vspace{3pt}\noindent \textbf{Retrieval.} 
This task involves cross-modal retrieval and disease retrieval. Cross-modal retrieval aims to retrieve the correct caption for a given image and vice versa. Disease retrieval, on the other hand, utilizes the disease names to retrieve captions or images with the same disease label, which is proposed to demonstrate the effectiveness of the knowledge encoder.
For cross-modal retrieval, we follow PLIP~\cite{huang2023leveraging} to split the ARCH~\cite{gamper2021multiple} dataset into Arch-PubMed and Arch-book. In addition, we also gather image-text pairs from publicly available educational resources and curate a retrieval dataset, termed as \textbf{PathPair}, consisting of 9,358 pathology image-caption pairs with known 1676 disease labels. For disease retrieval, we utilize the captions, images, and their disease labels in PathPair.

\vspace{3pt}\noindent \textbf{Zero-shot Patch Classification.} 
This task involves zero-shot classification on patch-level pathology images.
Specifically, at inference time, 
we randomly select one template from the 21 templates in CONCH~\cite{lu2023towards} and one type synonym from the corresponding name list (exhibited in Supplementary Materials) to yield a text prompt for each type, e.g. a histopathological image of CLASSNAME.~(template) + beast invasive carcinoma (synonym) $\rightarrow$ a histopathological image of beast invasive carcinoma. This process is repeated 100 times in every experiment.
Following PLIP~\cite{huang2023leveraging} and Quilt1M~\cite{ikezogwo2024quilt}, 
we adopt these patch-level pathology image datasets: BACH~\cite{aresta2019bach}, NCT-CRC-HE-100K~\cite{kather2018100}, KatherColon~\cite{kather2019predicting}, LC25000~\cite{borkowski2019lung}, RenalCell~\cite{brummer2022integrative}, 
SICAP~\cite{silva2020going}, SkinCancer~\cite{kriegsmann2022deep}, WSSS4LUAD~\cite{han2022wsss4luad}. 
Each dataset includes multiple types of H\&E stained cell micrographs.

\vspace{3pt}\noindent \textbf{Zero-shot WSI Tumor Subtyping.} 
This task involves zero-shot tumor subtyping on pathology whole slide images of common and rare cancers.
We follow MI-Zero~\cite{lu2023visual} and CONCH~\cite{lu2023towards} for evaluation.
Specifically, we first divide WSI into $256\times 256$ patches and then predict the class label of each image patch in a zero-shot manner.
The tumor type of the whole slide is then provided by integrating Top-K predictions on patches.
For this task, we also employ the templates and tumor synonyms from CONCH~\cite{lu2023towards} to randomly generate 100 text prompts (listed in Supplementary materials) for each tumor subtype.
(i) \textbf{For common cancers}, we follow existing research~\cite{lu2023visual,lu2023towards} and collect 525 WSIs with three tumor types, including 150 breast carcinoma (BRCA), 
150 non-small cell lung cancer (NSCLC) histopathology slides, 
and 225 renal cell carcinoma (RCC) slides, from TCGA. 
Specifically, BRCA consists of two subtypes: invasive ductal carcinoma (IDC) and invasive lobular carcinoma (IDC). NSCLC contains lung adenocarcinoma (LUAD) and lung squamous cell carcinoma (LUSC). RCC is divided into chromophobe renal cell carcinoma (CHRCC), clear-cell renal cell carcinoma (CCRCC), and papillary renal cell carcinoma (PRCC). Each tumor subtype has 75 WSIs.
(ii) \textbf{For rare cancers}, we collected 41 WSIs of rare subtypes for breast cancer (6 for Intraductal papillary adenocarcinoma with invasion, 6 for Medullary carcinoma, NOS, 13 for Metaplastic carcinoma, NOS 
and 16 for Mucinous adenocarcinoma), from TCGA and merge them with common breast cancers (75 for IDC and 75 for ILC).

\subsection{Evaluation Metrics.}
\noindent \textbf{Retrieval.} 
To evaluate the retrieval performance, we adopt the Recall@K metric, 
suggesting the ratio of correctly retrieved queries in Top-K retrieved samples.

\vspace{3pt}\noindent \textbf{Zero-shot Patch Classification.} 
For zero-shot classification tasks, we adopt the same metric as PLIP~\cite{huang2023leveraging}, namely, weighted F1 (wF1). 
We report the median, the first, and the third quartile (Q1, Q3) of wF1 across all text 100 prompts. 

\vspace{3pt}\noindent \textbf{Zero-shot WSI Tumor Subtyping.}
We adopt the same metric as MI-Zero~\cite{lu2023visual}, namely, balanced accuracy at Top-K pooling to measure the zero-shot tumor subtyping performance on WSIs. We also report the median, the first, and the third quartile (Q1, Q3) of the balanced accuracy across all 100 text prompts.

\subsection{Implementation Details}
\vspace{3pt}\noindent \textbf{Knowledge Encoder Pretraining.} 
We adopt the architecture of PubMedBERT~\cite{gu2021domain} to encode knowledge. The embedding dimension is set to 512. The temperature parameter $\tau$ is set to $0.04$ in Eq.~\ref{eq:2}. The batch size is set to 256, including 32 disease entities with 8 instances per entity. 

\vspace{3pt}\noindent \textbf{Pathology Image-text Pretraining.} To achieve a fair comparison with PLIP and Quilt1M, 
we utilize the same visual encoder~(ViT-B-32) and initialization of the visual encoder (CLIP~\cite{radford2021learning}), termed by KEP-32, and set the input image size to $224 \times 224$.  
Additionally, we also develop a KEP variant named KEP-16 that is initialized by the visual weights of BiomedCLIP~\cite{zhang2023large}~(ViT-B-16). The batch size and learning rate are set to 256 and 1e-5, respectively. The temperature in Eq.~\ref{eq:4} and Eq.~\ref{eq:5} is set to 0.04 across all experiments. For OpenPath, we conduct the pathology VLP for 30 epochs, While for Quilt1M, 15 epochs are adopted.

\section{Results}
In this section, we show the results on the downstream tasks, to evaluate the effectiveness of proposed knowledge-enhanced representation learning.
Note that for a fair comparison with the publicly available model released by the original authors, in all experiments, we separately train KEP on the OpenPath~\cite{huang2023leveraging} dataset, 
and Quilt1M~\cite{ikezogwo2024quilt} dataset,
and then report results on downstream tasks.

\subsection{Retrieval}
In this section, we evaluate the performance of retrieval,
including cross-modal retrieval on three datasets and disease retrieval on PathPair.

\begin{table}[t]
\centering
\caption{Performance comparison with PLIP and QuiltNet on three retrieval datasets. All models are pre-trained on the OpenPath and Quilt1M datasets, respectively. i2t and t2i denote image-to-text and text-to-image retrieval, respectively. Bold fonts suggest the best performance.}
\scalebox{0.71}{
\begin{tabular}{C{1.7cm}L{0.7cm}C{1.4cm}C{1.8cm}C{1.7cm}C{1.7cm}C{1.7cm}C{1.7cm}C{1.7cm}C{1.7cm}}
\toprule
Training &\multirow{2}{*}{Task} & \multirow{2}{*}{Model} & Knowledge & \multicolumn{2}{c}{Arch-PubMed} & \multicolumn{2}{c}{Arch-book} & \multicolumn{2}{c}{PathPair}\\ 
dataset & & & Enhancement & Recall@10 &Recall@50  & Recall@10 &Recall@50  & Recall@10 &Recall@50 \\ \midrule
\multirow{6}{*}{OpenPath} & \multirow{3}{*}{i2t} &PLIP & \textcolor{red}{\ding{55}} & 0.067 & 0.185 & 0.152 & 0.393 & 0.038  & 0.119 \\
& &KEP-32 & \textcolor{green}{\ding{51}} & 0.098 &0.283  & 0.164  &0.404  & 0.071  & 0.205\\ 
& &KEP-16 & \textcolor{green}{\ding{51}} & \textbf{0.163}  &\textbf{0.398}  &\textbf{0.244}  &\textbf{0.537} &\textbf{0.108} &\textbf{0.275} \\ \cmidrule{2-10} 
&\multirow{3}{*}{t2i} &PLIP & \textcolor{red}{\ding{55}} & 0.067 &0.181 &0.165 &0.419 &0.047 &0.133 \\
& &KEP-32 & \textcolor{green}{\ding{51}} &0.085 & 0.226 & 0.148 & 0.365 & 0.061 & 0.171 \\
& &KEP-16 & \textcolor{green}{\ding{51}} & \textbf{0.138} & \textbf{0.339} & \textbf{0.238} &\textbf{0.533} &\textbf{0.093} & \textbf{0.247} 
\\ \midrule

\multirow{6}{*}{Quilt1M} & \multirow{3}{*}{i2t} &QuiltNet & \textcolor{red}{\ding{55}} &0.139  &0.326  &0.188  &0.407 &0.065 &0.166 \\
& &KEP-32 & \textcolor{green}{\ding{51}} & 0.140 & 0.327 & 0.240 & 0.521 &  0.084 & 0.221\\
& &KEP-16 & \textcolor{green}{\ding{51}} & \textbf{0.196} & \textbf{0.421} & \textbf{0.282} & \textbf{0.564} & \textbf{0.108} & \textbf{0.254}
\\ \cmidrule{2-10} 
&\multirow{3}{*}{t2i} &QuiltNet & \textcolor{red}{\ding{55}} & 0.122 &0.293 &0.204 &0.429 &0.071 &0.195 \\
& &KEP-32 & \textcolor{green}{\ding{51}} &0.135 & 0.326 & 0.275 & 0.568 & 0.106 & 0.276 \\
& &KEP-16 & \textcolor{green}{\ding{51}} & \textbf{0.176} & \textbf{0.404} & \textbf{0.340} &\textbf{0.621} &\textbf{0.136} & \textbf{0.326} 
 \\ \bottomrule  
\end{tabular}
}
\label{tab:3}
\end{table}

\begin{table}
\centering
\caption{Performance comparison of disease retrieval. l2t and i2l denote label-to-text and image-to-label, respectively. Bold fonts suggest the best performance.}
\scalebox{0.77}{
\begin{tabular}{L{1.2cm}L{1.8cm}C{1.8cm}C{1.8cm}C{1.8cm}|C{1.8cm}C{1.8cm}C{1.8cm}}
\toprule
 \multirow{2}{*}{Task}&\multirow{2}{*}{Metrics} & \multicolumn{3}{c}{OpenPath} & \multicolumn{3}{c}{Quilt1M}
\\ 
 & & PLIP & KEP-32 &KEP-16 &QuiltNet  &KEP-32  & KEP-16 \\ \midrule
 \multirow{2}{*}{l2t} & Recall@10 &0.408 & 0.693 & \textbf{0.699} & 0.211 & 0.648 & 0.643\\ 
 & Recall@50 &0.536 & 0.832 & \textbf{0.835} & 0.325 & 0.812 & 0.808\\ \midrule
 \multirow{2}{*}{i2l} &Recall@10 & 0.114 & 0.162 & 0.223 & 0.135 & 0.226 & \textbf{0.259} \\ 
 & Recall@50 &0.292 & 0.381 & 0.478 & 0.322 & 0.473 & \textbf{0.519} \\
 \bottomrule   
\end{tabular}
}
\label{tab:5}
\end{table}

\vspace{3pt} \noindent \textbf{Cross-modal Retrieval.}
In Table~\ref{tab:3}, we demonstrate the results for different models pretrained on OpenPath and Quilt1M. It can be seen that KEP-32 pretrained on OpenPath outperforms PLIP on all datasets with respect to the image-to-text retrieval task, especially on Arch-PubMed and PathPair with more than 3\% boost on R10. As for the text-to-image task, KEP-32 also achieves better performance on the dataset of Arch-PubMed and PathPair. Furthermore, our KEP-16 model improves the retrieval performance by a large margin on all datasets across both retrieval tasks.
Similar results can be concluded that KEP-32 pretrained on Quilt1M improves the performance on all datasets for both retrieval tasks, which demonstrates the effectiveness of knowledge guidance for visual-language pretraining on cross-modal retrieval.

\vspace{3pt} \noindent \textbf{Disease Retrieval.}
Table~\ref{tab:5} shows the performance of disease retrieval on the PathPair dataset
for different models. 
It can be seen that although the scale of Quilt1M is 5 times OpenPath, 
models pretrained on OpenPath often outperforms that on Quilt1M for the label-to-text task. 
We conjecture this may be caused by the quality of captions in Quilt1M. 
Our approach KEP-32 outperforms PLIP and Quilt1M by a large margin on both label-to-text and image-to-label tasks, suggesting that the knowledge encoder contributes to paying attention to the key disease information when encoding captions, and thus improves the alignment between images and their disease labels.

\subsection{Zero-shot Patch Classification}
\begin{figure}[t]
    \centering
    \includegraphics[scale = 0.44]{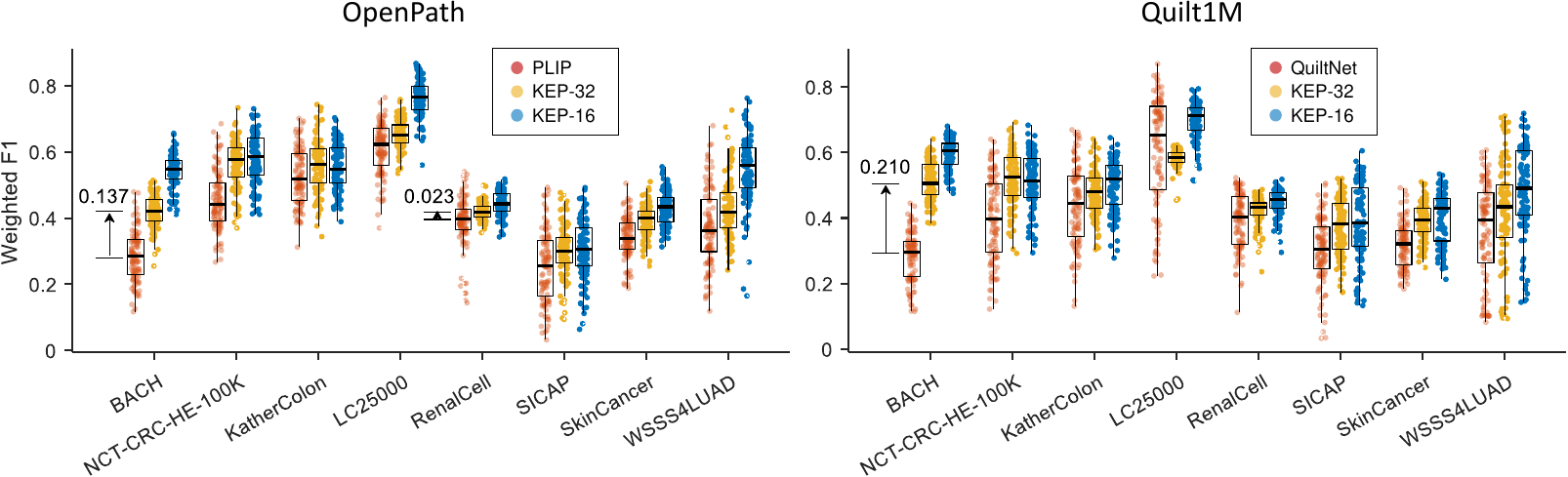}
    \caption{The comparison of zero-shot patch classification between different models. The left and the right subfigures suggest pretraining on OpenPath and Quilt1M, respectively. The visual encoders of KEP-32 and KEP-16 are initialized by CLIP (ViT-B-32) and BiomedCLIP (Vit-B-16), respectively. The number of points for every box is 100, with each representing the performance of one text prompt. The upper, center, and lower line of each box denote the first, median, and third quartile of the distribution.}
    \label{fig:result_zeroshot_all}
\end{figure}

In this section, we evaluate the performance of zero-shot classification on patch-level pathology images from 8 datasets.
We report the performance distribution (Fig.~\ref{fig:result_zeroshot_all}), where each point in the figure denotes the performance of one prompt.

\vspace{3pt} \noindent \textbf{Comparison to PLIP. } 
In Fig.~\ref{fig:result_zeroshot_all}~(left), we demonstrate the performance comparison with PLIP~\cite{huang2023leveraging}, 
by only pretraining on OpenPath image-text pairs. 
Compared with PLIP, KEP-32 achieves better zero-shot classification performance on all datasets, with a maximum enhancement of 0.137 on the BACH dataset. 
Moreover, with the medical-specific initialization of the visual encoder~(BiomedCLIP),
KEP-16 can further improve the performance significantly in most datasets. 
Note that, the boxes of KEP variants are generally shorter than that of PLIP, suggesting that our approach is less sensitive to the varying text prompts than PLIP, which can be demonstrated by the visualization of prompt embeddings, shown in Supplementary Materials.

\vspace{3pt} \noindent \textbf{Comparison to QuiltNet.} 
As shown in Fig.~\ref{fig:result_zeroshot_all}~(right), 
we exhibit the performance comparison between KEP and QuiltNet~\cite{ikezogwo2024quilt} pretrained on Quilt1M dataset. 
Similar to results on OpenPath, KEP-32 outperforms Quilt1M in seven out of eight datasets, especially for the BACH, KatherColon, and SkinCancer datasets. KEP-16 can further improve the weighted F1 score. 
Most boxes of KEP variants are also shorter than that of Quilt1M. 
The visualization of prompt embeddings, 
shown in Supplementary Materials, 
maintains that pathology knowledge can reduce the ambiguities of image-text alignment and thus improve the zero-shot classification performance.

\begin{figure}[t]
    \centering
    \includegraphics[scale = 0.45]{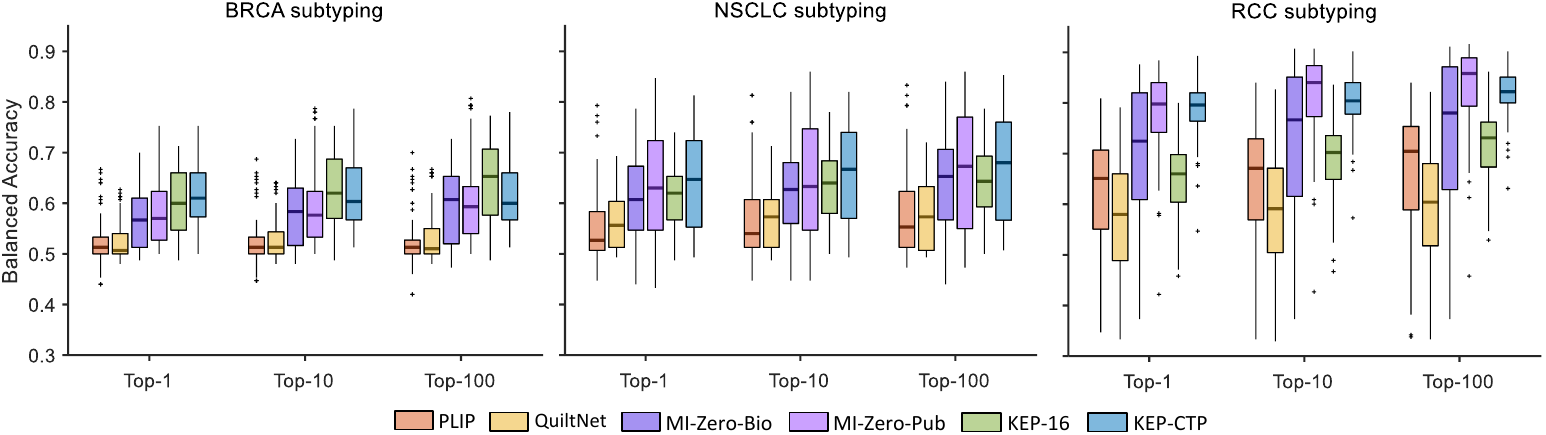}
    \caption{The performance comparison of tumor subtyping on TCGA-BRCA (common), TCGA-NSCLC, and TCGA-RCC WSIs. The upper, center and lower line of each box denote the first, median, and third quartile of the performance distribution, respectively. The scattered points represent outliers. KEP-16 and KEP-CTP are trained on OpenPath with the visual encoder initialized by BiomedCLIP~\cite{zhang2023large} and CTransPath~\cite{wang2022transformer}, respectively. MI-Zero-Bio and MI-Zero-Pub are two variants of MI-Zero with different initialization of text encoder. Their visual encoders are initialized by CTransPath. In addition, MI-Zero adopts in-house pathology reports to pretrain their text encoders.}
    \label{fig:result_wsi_all}
\end{figure}

\begin{table}
\centering
\caption{Performance comparison of tumor subtyping on six BRCA subtypes, including 2 common and 4 rare cancers. Bold fonts suggest the best performance.}
\label{tab:rare}

\scriptsize
\setlength{\tabcolsep}{6pt}
\scalebox{0.95}{
\begin{tabular}{lcccccc}
\toprule
  \multirow{2}{*}{Model}&\multirow{2}{*}{Training dataset} &\multicolumn{2}{c}{Top-1} & \multicolumn{2}{c}{Top-10}
\\ 
 & & Common & Rare & Common & Rare  \\ \midrule
 PLIP & OpenPath & 0.113 & 0.115 &0.119 &0.119  \\
 QuiltNet &Quilt1M & 0.278 & 0.094 & 0.279 &0.102  \\ 
 \rowcolor[RGB]{240,240,240} KEP-16 &OpenPath & \textbf{0.343} &\textbf{0.217} &\textbf{0.368} &\textbf{0.245}  \\  \midrule
 MI-Zero-Bio &In-house \& Web data & 0.390 & 0.214 & 0.403 & 0.237  \\
 MI-Zero-Pub &In-house \& Web data &0.331 &\textbf{0.282} & 0.347 & \textbf{0.304}  \\
 \rowcolor[RGB]{240,240,240} KEP-CTP &OpenPath &\textbf{0.443} & \textbf{0.282} & \textbf{0.432} & 0.301 \\
 \bottomrule
\end{tabular}
}
\end{table}

\subsection{Zero-shot WSI Tumor Subtyping}
In this section, we evaluate the transfer ability of different models for tumor subtyping on common and rare cancers. 

\vspace{3pt}\noindent\textbf{Common cancers.}
Fig.~\ref{fig:result_wsi_all} shows the performance comparison of tumor subtyping on common cancers, including TCGA-BRCA (common), TCGA-NSCLC and TCGA-RCC WSIs. 
MI-Zero-Bio and MI-Zero-Pub are two variants of MI-Zero with different text encoders. The visual encoders of the two MI-Zero variants are initialized by CTransPath~\cite{wang2022transformer}. 
To achieve a fair comparison, we also train a KEP variant named KEP-CTP with the visual encoder initialized by CTransPath. Both KEP-16 and KEP-CTP are trained on the OpenPath dataset. It can be seen that our approach KEP-16 and KEP-CTP outperform PLIP and QuiltNet on all datasets.
Note that, the comparisons between variants of MI-Zero and KEP-CTP are actually not fair for our method, as the MI-Zero variants have also been pretrained on massive in-house data -- over 550k pathology reports from hospitals~\cite{lu2023visual}.
Yet, KEP-CTP still outperforms MI-Zero-Bio across all datasets and achieves comparable performance with MI-Zero-Pub on NSCLC and RCC.

\vspace{3pt}\noindent\textbf{Rare cancers.} 
Table~\ref{tab:rare} exhibits the median balanced accuracy of tumor subtyping on common and rare breast cancers. It can be seen that our method KEP-16 pretrained on OpenPath outperforms PLIP and QuiltNet by a large margin on both common and rare breast cancers. 
Moreover, KEP-CTP significantly outperforms MI-Zero variants on common breast cancers while achieving comparable performance on the rare ones.

\subsection{Ablation Study} 

In this section, we explore the impact of model architecture, initialization, and other hyperparameters, as shown in Table~\ref{tab:6} and Table S3 in Supplementary Materials. 
Comparing the performance between different models, 
we can draw the following observations: 
(i) visual projection head can facilitate the zero-shot classification performance; 
(ii) our pretrained knowledge encoder can enhance the performance across all tasks; (iii) the frozen knowledge branch can further improve the performance of zero-shot classification and disease retrieval tasks while the frozen PubmedBERT can not; (iv) a better initialization of visual encoder contributes to significant performance improvement on all tasks. (v) the Adasp loss is better than the Triplet loss for the knowledge encoding.

\begin{table}[t]
\centering
\caption{Experimental results of ablation study on model architecture and initialization. Arch-PubMed and PathPair suggest the text-to-image retrieval and the disease retrieval task, respectively. PMB and KB denote PubMedBERT and our pretrained pathology knowledge encoder, respectively. BCLIP denotes BiomedCLIP. Bold fonts and underline suggest the best and the second-best performance, respectively.}

\label{tab:6}
\footnotesize
\scalebox{0.8}{
\begin{tabular}{L{1.2cm}C{1.cm}C{1.5cm}C{1.4cm}C{1.4cm}C{3.3cm}C{1.cm}C{1.cm}C{1.cm}C{1.cm}}
\toprule
 Visual & Text & Projection  & Knowledge & Metric & KatherColon & \multicolumn{2}{c}{Arch-PubMed} & \multicolumn{2}{c}{PathPair} \\
  Init.  & Init.  & head   &  distill & loss  & Median (Q1, Q3) &R10  &R50  &R1  &R5\\ \midrule
  Scratch & PMB & & & & 0.322 (0.272, 0.392) & 0.037 & 0.121 & 0.056 & 0.113 \\
  CLIP & PMB & & & & 0.407 (0.358, 0.452) & 0.081 & 0.214 & 0.161 & 0.316\\ 
  CLIP & PMB & \checkmark & & & 0.486 (0.448, 0.517) & 0.073 & 0.210 & 0.161 & 0.305\\ 
  BCLIP & PMB & \checkmark  & & & 0.553 (0.503, 0.591) & \underline{0.127} & \textbf{0.347} & 0.187 & 0.350\\
  CLIP & PMB & \checkmark & \checkmark (PMB) &  & 0.484 (0.435, 0.520) & 0.063 & 0.182 & 0.034 & 0.089\\ 
 CLIP & KB & \checkmark  & & Adasp& 0.530 (0.477, 0.600) & 0.086 & 0.232 & 0.270 & 0.482\\
 CLIP & KB & \checkmark  & \checkmark (KB) & Adasp & \underline{0.563} (\underline{0.505}, \underline{0.610}) & 0.085 & 0.226 &\underline{0.409} & \textbf{0.618}\\
 CLIP & KB & \checkmark  & \checkmark (KB) & Triplet & 0.531 (0.482, 0.580) & 0.087& 0.227 & 0.265 & 0.493\\
 BCLIP & KB & \checkmark  & \checkmark (KB) & Adasp & \textbf{0.580} (\textbf{0.517}, \textbf{0.631}) & \textbf{0.138} & \underline{0.339} &\textbf{0.424} & \textbf{0.618}\\
 \bottomrule
\end{tabular}
}
\end{table}

\section{Conclusion}
In this paper, we address the problem of knowledge-enhanced visual-language pretraining on computational pathology. We first curate a pathology knowledge tree that integrates the informative attributes of diseases requiring pathological diagnosis. We then propose a novel knowledge encoding approach based on metric learning to model structured pathological knowledge. With the guidance of the pretrained knowledge encoder, we conduct extensive visual-language pretraining on pathology image-caption pairs. To demonstrate the effectiveness of our approach, we evaluate pretrained VLP models on three downstream tasks, including retrieval, zero-shot classification on patch-level pathology images, and zero-shot tumor subtyping on pathology WSIs. Quantitative experimental results demonstrate that pathology knowledge can significantly improve the performance across different tasks.

\section*{Acknowledgements}
This work is supported by the National Key R\&D Program of China (No. 2022ZD0160702) and China Postdoctoral Science Foundation (Certificate Number: 2023M741850).

%
%
\bibliographystyle{splncs04}
\bibliography{main}

\clearpage
\setcounter{page}{1}

\onecolumn
\let\titleold\title
\renewcommand{\title}[1]{\titleold{#1}\newcommand{\thetitle}{#1}}
{
    \centering
    \Large
    \vspace{1.5em}\textbf{Supplementary Material} \\
    \vspace{1.0em}
}

\setcounter{figure}{0}
\setcounter{table}{0}
\setcounter{equation}{0}
\setcounter{section}{0}

\renewcommand*{\thefigure}{S\arabic{figure}}
\renewcommand*{\thetable}{S\arabic{table}}
\renewcommand*{\theequation}{S\arabic{equation}}

\noindent This Supplementary Material contains the following parts:

\vspace{5mm}
\noindent \textbf{1. Dataset Statistics} 

The summary of datasets used in this paper.

\vspace{5mm}
\noindent \textbf{2. Additional Results on Zero-shot WSI Tumor Subtyping} 

Additional results on zero-shot WSI tumor subtyping for KEP variants pretrained on Quilt1M and KEP variants with CNN-based visual encoders.

\vspace{5mm}
\noindent \textbf{3. Additional Ablation Results} 

Additional ablations on the weight parameter $\alpha$ in the loss function.

\vspace{5mm}
\noindent \textbf{4. Discussion} 

Details and discussion about the soft version of max-min positive similarity, frozen knowledge encoder, and robustness towards text prompts.

\vspace{5mm}
\noindent \textbf{5. Text Prompts} 

The list of text prompts used in this paper.

\clearpage
\section{Dataset Statistics}

\begin{figure}
    \centering
    \includegraphics[scale = 0.95]{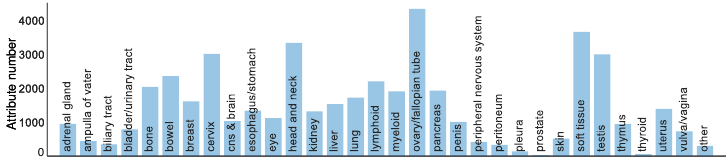}
    \caption{The disease distribution in PathKT.}
    \label{fig:pathkt_statistics}
\end{figure}

\begin{table}
\centering
\caption{Statistics of the dataset for knowledge encoder pretraining. Syn., Def., His. and Cyt. denote synonyms, definitions, histological and cytological features.}

\scalebox{0.8}{
\begin{tabular}{L{3cm}C{2cm}C{2cm}C{2cm}C{2cm}C{2cm}C{2cm}C{2cm}}
\toprule
Attributes    &Syn.    &Def.        & His.       & Cyt. & All \\
\midrule
\# Tissue      &32    &32        & 30       & 30  & 32\\ 
\# Disease entity  & 3,348   & 4,661   & 3,558   & 1,117  & 4,718   \\
\# Attributes  & 30,950   & 10,097   & 3,558   & 1,117  & 50,470 \\   \bottomrule   
\end{tabular}
}
\label{tab:1}
\end{table}

\begin{table}[h!]
  \centering
  \caption{Statistics of test datasets. Classification and subtyping suggest zero-shot patch-level image classification and WSI tumor subtyping, respectively.}
  \label{tab:2}
   \scalebox{0.8}{
  \begin{tabular}{L{2.8cm}|L{3.8cm}L{2.6cm}L{2.6cm}L{2.6cm}}
  \toprule
   Task  &Dataset  &\#images  &\#types  &tissue    \\ \midrule
  
    \multirow{3}{*}{Retrieval} &Arch-PubMed &1923 &--  &Multiple\\
    &Arch-book &1306 &--  &Multiple\\
    &PathPair &9,358 &1676  &Multiple\\
    \midrule
    \multirow{8}{*}{Classificaiton} &BACH &400 &4  &Breast\\
    &NCT-CRC-HE-100K &100K &9   &Colorectal \\
    &KatherColon &7,180 &9   &Colon \\
    &LC25000 &25K &5 &Lung, Colon \\ 
    &RenalCell &36,687 &5 &Renal \\ 
    &SICAP &12,081 &4 &Prostate \\ 
    &SkinCancer &129,369 &16 &Skin \\ 
    &WSSS4LUAD &4,693 &3 &Lung \\ 
    \midrule
    \multirow{3}{*}{Subtyping} &TCGA-BRCA-common &150 &2  &Breast\\
    &TCGA-BRCA-rare &41 &4  &Breast\\
    &TCGA-NSCLC &150 &2  &Lung\\
    &TCGA-RCC &225 &3  &Renal\\
  \bottomrule
  \end{tabular}
  }
\end{table}

\clearpage
\section{Additional Results on Zero-shot WSI Tumor Subtyping}

\begin{figure}[!h]
    \centering
    \includegraphics[scale = 0.53]{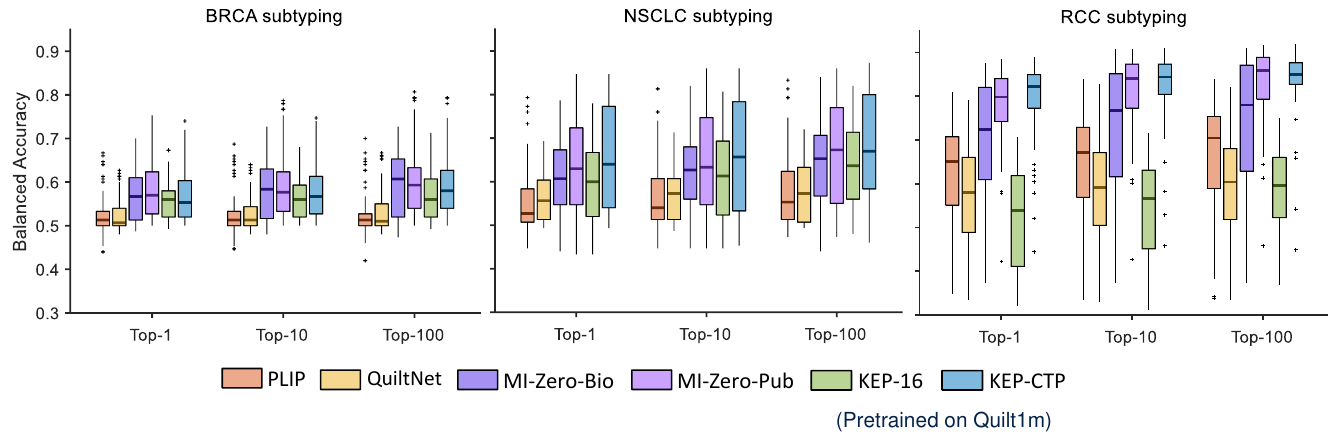}
    \caption{The performance comparison of tumor subtyping on TCGA-BRCA, TCGA-NSCLC, and TCGA-RCC WSIs. The upper, center and lower line of each box denote the first, median, and third quartile of the performance distribution, respectively. The scattered points represent outliers. 
    KEP-16 and KEP-CTP are trained on Quilt1M with the visual encoder initialized by BiomedCLIP~\cite{zhang2023large} and CTransPath~\cite{wang2022transformer}, respectively.}
    \label{fig:result_quit1m_wsi_all}
\end{figure}

In this section, we provide additional results on zero-shot WSI tumor subtyping for the KEP variants that are pretrained on Quilt1M, as shown in Fig.~\ref{fig:result_quit1m_wsi_all}. It can be seen that KEP-16 achieves better performance than QuiltNet on BRCA and NSCLC datasets. KEP-CTP can further improve the performance, especially on the RCC dataset.

\begin{figure}[!h]
    \centering
    \vspace{-2mm}
    \includegraphics[scale = 0.45]{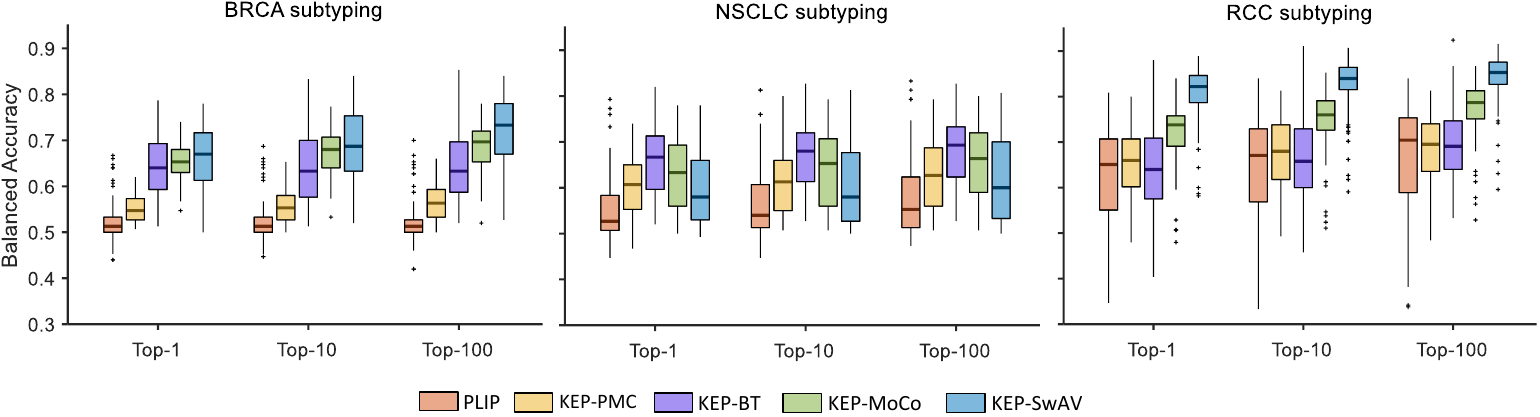}
    \caption{The performance comparison of tumor subtyping on TCGA-BRCA, TCGA-NSCLC, and TCGA-RCC WSIs. The upper, center and lower line of each box denote the first, median, and third quartile of the performance distribution, respectively. The scattered points represent outliers. 
    KEP-PMC is trained on OpenPath with the visual encoder initialized by PMC-CLIP~\cite{lin2023pmc}, KEP-BT, KEP-MoCo, and KEP-SwAV are trained on OpenPath with the visual encoder initialized by PathSSL-BT, PathSSL-MoCo, and PathSSL-SwAV~\cite{kang2023benchmarking}, respectively.}
    \label{fig:result_plip_wsi_ssl_pmc}
\end{figure}

Furthermore, we change the backbone of KEP's visual encoder to convolutional neural networks and initialize it with PMC-CLIP~\cite{lin2023pmc} and PathSSL~\cite{kang2023benchmarking}. 
PMC-CLIP refers to a visual-language model pretrained on medical image-text pairs crawled from PubMed papers. PathSSL consists of PathSSL-BT, PathSSL-MoCo, and PathSSL-SwAV, which are self-supervised models pretrained on TCGA pathology images with different SSL strategies, including Barlow Twins~\cite{zbontar2021barlow}, MoCo v2~\cite{chen2020improved}, and SwAV~\cite{caron2020unsupervised}. Except for the visual encoder, all training details are consistent with the main text. 

Fig.~\ref{fig:result_plip_wsi_ssl_pmc} shows the performance comparison between PLIP and different KEP variants pretrained on OpenPath. It can be seen that almost all KEP variants outperform PLIP on all datasets, suggesting that our pretraining approach can seamlessly generalize to CNN-based visual encoders with the same parameter configuration. In particular, KEP-MoCo and KEP-SwAV significantly improve the subtyping performance on breast and renal tumors.

 \section{Additional Ablation Results}
Table~\ref{tab:add_ablation} shows that KEP achieves the best performance on KatherColon dataset when the loss weight $\alpha$ is set to 0.3.

 \begin{table}[h!]
\centering
\caption{Experimental results of the ablation study on the loss weight $\alpha$. Bold fonts suggest the best performance.}
\label{tab:add_ablation}
\footnotesize
\scalebox{0.8}{
\begin{tabular}{L{1.5cm}C{1.5cm}C{1.5cm}C{1.5cm}C{1.5cm}C{1.5cm}C{1.5cm}C{1.5cm}}
\toprule
$\alpha$ &0.01  & 0.05 & 0.1 & 0.3 &0.5  &0.7  &0.9  \\ \midrule
Median  &0.531 &0.524&0.531 & \textbf{0.563} & 0.556 & 0.496 & 0.481   \\
Q1   &0.478 &0.480&0.472& \textbf{0.505} & 0.497 & 0.447 & 0.428    \\
Q3   &0.589 &0.587&0.571& \textbf{0.610} & 0.595 & 0.540 & 0.533    \\ \bottomrule   
\end{tabular}
}
\end{table}

\section{Discussion}

\vspace{3pt}\noindent\textbf{Soft Version of Max-min Positive Similarity.}
\begin{equation}\label{eqs1}
\small
    \begin{aligned}
    S_i^{+} = \max_{p}{\min_{q}{\left< \mathbf{z}_p^i, \mathbf{z}_q^i \right> }} &\approx \tau \log \left( \sum_{p=1}^k{e^{\frac{\min_{q}{\left< \mathbf{z}_p^i, \mathbf{z}_q^i \right> }}{\tau}}} \right)
    \end{aligned}
\end{equation}
\begin{equation} \label{eqs2}
\small
     \min_{q}{\left< \mathbf{z}_p^i, \mathbf{z}_q^i \right> } \approx -\tau \log \left( \sum_{q=1}^k{e^{-\frac{\left< \mathbf{z}_p^i, \mathbf{z}_q^i \right>}{\tau}}} \right) 
\end{equation}
\begin{equation}\label{eqs3}
\small
    S_i^{+} \approx \tau \log \left( \sum_{p=1}^k{\frac{1}{\sum_{q=1}^k{e^{-\frac{\left< \mathbf{z}_p^i, \mathbf{z}_q^i \right>}{\tau}}}}} \right)
\end{equation}

\vspace{3pt}\noindent\textbf{Frozen Knowledge Encoder.}
Here, we discuss the advantages of the frozen knowledge encoder. Although the visual-language pretraining can start in a well-defined pathology embedding space curated by the knowledge initialization of the text encoder, the free text from the training pairs may impair the knowledge structure built from the pathology knowledge tree, leading to overfitting on the training image-text pairs. 
As a consequence, the frozen knowledge encoder across the entire training period holds two advantages: 
(i) the text embedding from this branch acts as a frozen knowledge continuously distilling to the active text encoder, which, therefore, can substantially keep the entire alignment procedure in the well-structured knowledge embedding space; 
(ii) the additional contrastive loss between the active text encoder and this branch serves as a regularization term to prevent the overfitting problem.

\begin{figure}
    \centering
    \includegraphics[scale = 0.42]{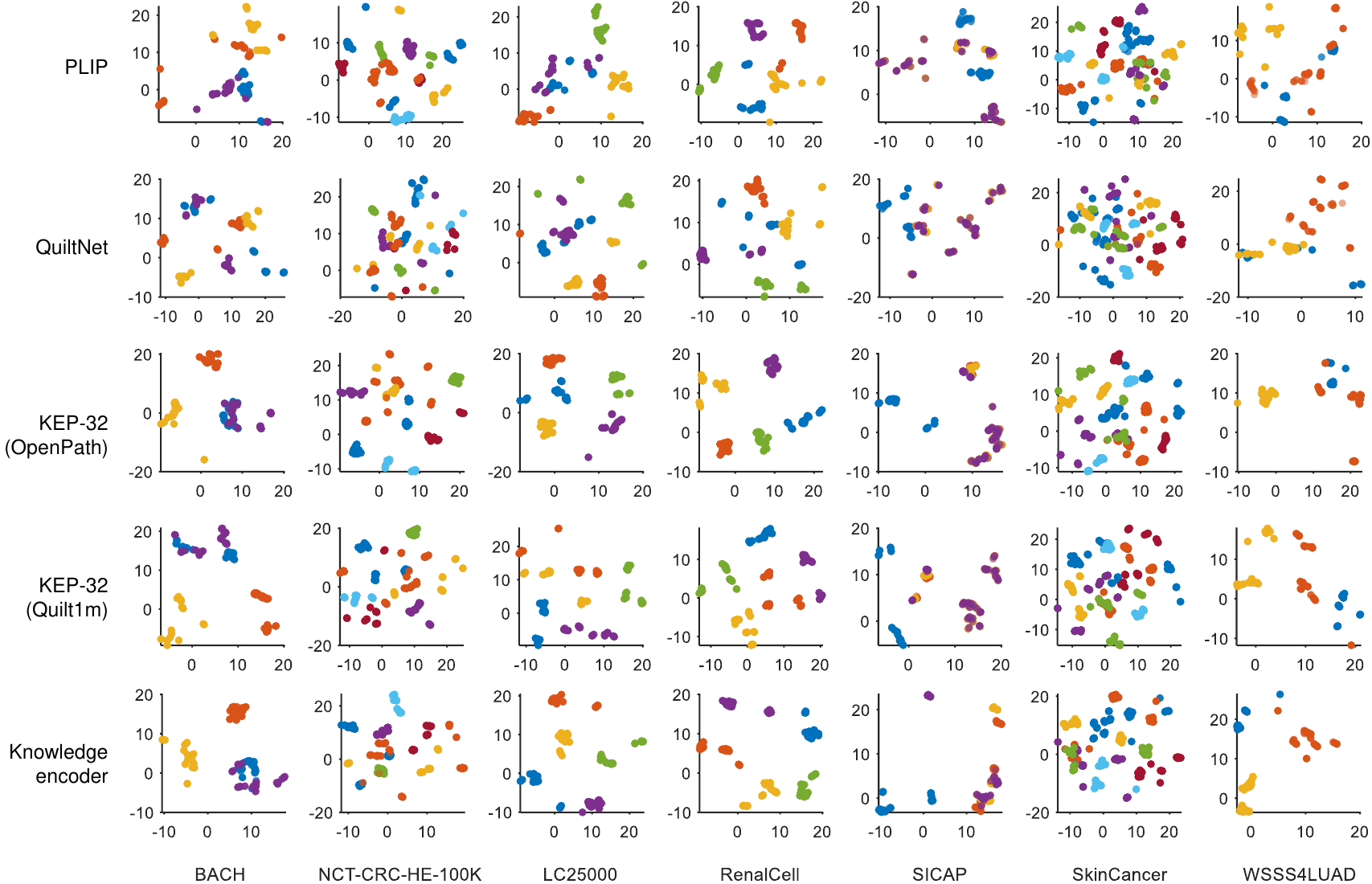}
    \caption{The UMAP visualization of prompt embeddings on different datasets. Each color represents a class of the dataset and consists of 100 points corresponding to the random 100 prompts for each category. NCT-CRC-HE-100K and KatherColon share the same text prompts. KEP-32 (OpenPath) and KEP-32 (Quilt1M) denote the prompt embeddings generated from the text encoder pretrained on the dataset of OpenPath and Quilt1M, respectively.
    Knowledge encoder denotes the prompt embeddings from the frozen knowledge encoder.}
    \label{fig:result_plip_umap}
\end{figure}

\vspace{3pt}\noindent\textbf{Robustness towards Text Prompts.}
To show the robustness of our approach towards different text prompts, 
we use UMAP~\cite{mcinnes2018umap} to visualize the prompt embeddings for each class, as shown in Fig.~\ref{fig:result_plip_umap}. 
It can be concluded that the prompt embeddings of PLIP and QuiltNet suffer from high similarities between different classes, especially in the dataset of BACH, LC25000, SICAP, and WSSS4LUAD, 
which therefore causes large ambiguities for zero-shot classification. 
In contrast, the embeddings generated by our proposed frozen knowledge encoder~(denoted as Knowledge encoder), are well-separated in most datasets, 
With the knowledge guidance, the text encoder of KEP, denoted by KEP-32, 
also outputs well-separated prompt embeddings of different classes, 
suggesting that pathology knowledge can substantially improve the structure of text embedding space.

\section{Text Prompts}
In this section, we list all text prompts used in this paper. The templates are shown in Tab.~\ref{tab:templates}. The synonyms for the patch-level dataset are shown in Tab.~\ref{tab:bach_synonyms}-\ref{tab:skincancer_synonyms}. The class names for TCGA WSIs are exhibited in Tab.~\ref{tab:brca_synonyms}-\ref{tab:rcc_synonyms}.

\vspace{0.2cm}
\begin{table}[!htb]
\centering
\caption{Prompt templates used in this paper, which are consistent with CONCH~\cite{lu2023towards},  CLASSNAME is replaced by the names/synonyms of classes.}
\footnotesize
\scalebox{0.8}{
\begin{tabular}{l}
\toprule
 CLASSNAME.\\
 a photomicrograph showing CLASSNAME.\\
 a photomicrograph of CLASSNAME.\\
 an image of CLASSNAME.\\
 an image showing CLASSNAME.\\
 an example of CLASSNAME.\\
 CLASSNAME is shown. \\
 this is CLASSNAME. \\
 there is CLASSNAME.\\
 a histopathological image showing CLASSNAME.\\
 a histopathological image of CLASSNAME.\\
 a histopathological photograph of CLASSNAME.\\
 a histopathological photograph showing CLASSNAME.\\
 shows CLASSNAME.\\
 presence of CLASSNAME.\\
 CLASSNAME is present.\\
 an H\&E stained image of CLASSNAME.\\
 an H\&E stained image showing CLASSNAME.\\
 an H\&E image showing CLASSNAME.\\
 an H\&E image of CLASSNAME.\\
 CLASSNAME, H\&E stain.\\
 CLASSNAME, H\&E.\\
 \bottomrule   
\end{tabular}
}
\label{tab:templates}
\end{table}

\begin{table}[!htb]
\centering
\caption{Class names of BACH.}
\footnotesize
\scalebox{0.8}{
\begin{tabular}{lp{12cm}}
\toprule
Class & Names/Synonyms\\
\midrule
Benign &breast non-malignant benign tissue; breast benign tissue; non-malignant benign tissue of breast\\
\midrule
InSitu &breast malignant in-situ carcinoma; breast in-situ carcinoma; malignant carcinoma in-situ of breast\\

\midrule
Invasive & breast malignant invasive carcinoma; breast invasive carcinoma; invasive carcinoma of breast\\

\midrule
Normal &normal breast tissue; breast normal tissue; breast non-cancerous tissue\\
 \bottomrule   
\end{tabular}
}
\label{tab:bach_synonyms}
\end{table}

\begin{table}[!htb]
\centering
\caption{Class names of NCT-CRC-HE-100K and KatherColon from CONCH~\cite{lu2023towards}.}
\footnotesize
\scalebox{0.8}{
\begin{tabular}{lp{12cm}}
\toprule
Class & Names/Synonyms\\
\midrule
ADI & adipose; adipose tissue; adipocytes; fat; fat cells\\
\midrule
BACK &background; penmarking; empty space; background artifacts\\

\midrule
DEB &debris; colorectal adenocarcinoma debris and necrosis; necrosis; necrotic debris\\

\midrule
LYM &lymphocytes; lymphoid aggregate; immune cells; lymphoid infiltrate; inflammatory cells\\

\midrule
MUC&mucus; mucin; mucus pool; mucin pool\\

\midrule
MUS&smooth muscle; smooth muscle tissue; muscle; muscularis propria; muscularis mucosa\\

\midrule
NORM&normal colon mucosa; uninvolved colon mucosa; benign colon mucosa; benign epithelium\\

\midrule
STR&cancer-associated stroma; tumor-associated stroma; stromal cells; stromal tissue; stroma\\

\midrule
TUM&colorectal adenocarcinoma epithelium; colorectal adenocarcinoma; tumor; adenocarcinoma; malignant epithelium\\

 \bottomrule   
\end{tabular}
}
\label{tab:kather_synonyms}
\end{table}

\vspace{1cm}

\begin{table}
\centering
\caption{Class names of LC25000.}
\footnotesize
\scalebox{0.8}{
\begin{tabular}{lp{11cm}}
\toprule
Class & Names/Synonyms\\
\midrule
lung\_aca &lung adenocarcinoma; adenocarcinoma of the lung; lung cancer, adenocarcinoma\\
\midrule
lung\_n &benign lung; benign lung tissues; non-malignant lung tissue\\

\midrule
lung\_scc &lung squamous cell carcinoma; squamous-cell carcinoma of the lung; squamous cell lung cancer\\

\midrule
colon\_aca &colon adenocarcinoma; adenocarcinoma of the colon; colon cancer, adenocarcinoma\\

\midrule
colon\_n &benign colon; benign colonic tissue; non-malignant colon tissue\\

 \bottomrule   
\end{tabular}
}
\label{tab:lc25000_synonyms}
\end{table}

\begin{table}
\centering
\caption{Class names of RenalCell.}
\scalebox{0.8}{
\begin{tabular}{lp{12cm}}
\toprule
Class & Names/Synonyms\\
\midrule
blood &red blood cells; red blood corpuscles; red cells; erythroid cells\\
\midrule
cancer &non-tumor; normal tissue; non-cancerous tissue\\

\midrule
normal &renal cancer; renal tumor; renal neoplasm; renal carcinoma\\

\midrule
other &torn adipose necrotic tissue; torn adipose tissue, necrosis; adipose necrotic tissue\\

\midrule
stroma &muscle fibrous stroma blood vessels; blood vessels, muscle fibrous stroma; muscle fibers and blood vessels in stroma\\

 \bottomrule   
\end{tabular}
}
\label{tab:renalcell_synonyms}
\end{table}

\begin{table}
\centering
\caption{Class names of SICAP from CONCH~\cite{lu2023towards}.}
\footnotesize
\scalebox{0.8}{
\begin{tabular}{lp{12cm}}
\toprule
Class & Names/Synonyms\\
\midrule
NC &non-cancerous tissue; non-cancerous prostate tissue; benign tissue; benign glands; benign prostate tissue; benign prostate glands\\
\midrule
G3 &gleason grade 3; gleason pattern 3; prostate cancer, gleason grade 3; prostate cancer, gleason pattern 3; prostate adenocarcinoma, well-differentiated; well-differentiated prostatic adenocarcinoma\\

\midrule
G4 &gleason grade 4; gleason pattern 4; prostate cancer, gleason grade 4; prostate cancer, gleason pattern 4; prostate adenocarcinoma, moderately differentiated; moderately differentiated prostatic adenocarcinoma\\

\midrule
G5 &gleason grade 5; gleason pattern 5; prostate cancer, gleason grade 5; prostate cancer, gleason pattern 5; prostate adenocarcinoma, poorly differentiated; poorly differentiated prostatic adenocarcinoma\\

\midrule
Tumor &prostatic adenocarcinoma; adenocarcinoma; prostate cancer; tumor tissue; cancerous tissue\\

 \bottomrule   
\end{tabular}
}
\label{tab:sicap_synonyms}
\end{table}

\begin{table}
\centering
\caption{Class names of WSSS4LUAD from CONCH~\cite{lu2023towards}.}
\footnotesize
\scalebox{0.8}{
\begin{tabular}{lp{12cm}}
\toprule
Class & Names/Synonyms\\
\midrule
normal &non-tumor; normal tissue; non-cancerous tissue\\
\midrule
stroma &tumor-associated stroma; cancer-associated stroma; tumor-associated stromal tissue; cancer-associated stromal tissue\\

\midrule
tumor &tumor tissue; tumor epithelial tissue; cancerous tissue\\

 \bottomrule   
\end{tabular}
}
\label{tab:wsss4luad_synonyms}
\end{table}

\begin{table}
\centering
\caption{Class names of SkinCancer.}
\footnotesize
\scalebox{0.8}{
\begin{tabular}{lp{10cm}}
\toprule
Class & Names/Synonyms\\
\midrule
necrosis &necrosis; necrotic tissue; necrotic cells\\
\midrule
skeletal &skeletal muscle; skeletal muscle cells; skeletal muscle tissue\\

\midrule
sweatglands &eccrine sweat glands; merocrine glands; skin eccrine sweat glands\\

\midrule
vessel &vessels; blood vessels; vessel\\

\midrule
elastosis &elastosis; elastosis of skin; skin elastosis\\
\midrule
chondraltissue &chondral tissue; chondral tissue of skin; skin chondral tissue\\
\midrule
hairfollicle &hair follicle; hair follicle of skin; skin hair follicle\\
\midrule
epidermis &epidermis; skin epidermis; epidermal cells\\
\midrule
nerves &nerves; nerve fibers; nerve axons\\
\midrule
subcutis &subcutis; subcutaneous tissue; skin subcutis; hypodermis; hypoderm\\
\midrule
dermis &dermis; skin dermis; corium; skin corium\\
\midrule
sebaceousglands &sebaceous; sebaceous gland; skin sebaceous\\
\midrule
sqcc &squamous-cell carcinoma; cutaneous squamous-cell carcinoma; squamous-cell carcinoma of the skin; squamous-cell skin cancer\\
\midrule
melanoma &melanoma in-situ; malignant melanoma; cutaneous melanoma\\
\midrule
bcc &basal-cell carcinoma; basal-cell cancer; basal-cell tumor\\
\midrule
naevus &naevus; mole; skin nevus\\

 \bottomrule   
\end{tabular}
}
\label{tab:skincancer_synonyms}
\end{table}

\begin{table}
\centering
\caption{Class names of TCGA-BRCA, which are consistent with CONCH~\cite{lu2023towards}.}
\footnotesize
\scalebox{0.8}{
\begin{tabular}{lp{12cm}}
\toprule
Class & Names/synonyms\\
\midrule
IDC & invasive ductal carcinoma; breast invasive ductal carcinoma; invasive ductal carcinoma of the breast; invasive carcinoma of the breast, ductal pattern; breast IDC\\
\midrule
ILC & invasive lobular carcinoma; breast invasive lobular carcinoma; invasive lobular carcinoma of the breast; invasive carcinoma of the breast, lobular pattern; breast ILC\\

 \bottomrule   
\end{tabular}
}
\label{tab:brca_synonyms}
\end{table}

\begin{table}
\centering
\caption{Class names of TCGA-NSCLC, which are consistent with CONCH~\cite{lu2023towards}.}
\footnotesize
\scalebox{0.8}{
\begin{tabular}{lp{12cm}}
\toprule
Class & Names/synonyms\\
\midrule
LUAD & adenocarcinoma; lung adenocarcinoma; adenocarcinoma of the lung; LUAD\\
\midrule
LUSC & squamous cell carcinoma; lung squamous cell carcinoma; squamous cell carcinoma of the lung; LUSC\\

 \bottomrule   
\end{tabular}
}
\label{tab:nsclc_synonyms}
\end{table}

\begin{table}
\centering
\caption{Class names of TCGA-RCC, which are consistent with CONCH~\cite{lu2023towards}.}
\footnotesize
\scalebox{0.8}{
\begin{tabular}{lp{11cm}}
\toprule
Class & Names/synonyms\\
\midrule

CCRCC &clear cell renal cell carcinoma; renal cell carcinoma, clear cell type; renal cell carcinoma of the clear cell type; clear cell RCC\\
\midrule
PRCC &papillary renal cell carcinoma; renal cell carcinoma, papillary type; renal cell carcinoma of the papillary type; papillary RCC\\
\midrule
CHRCC &chromophobe renal cell carcinoma; renal cell carcinoma, chromophobe type; renal cell carcinoma of the chromophobe type; chromophobe RCC\\

 \bottomrule   
\end{tabular}
}
\label{tab:rcc_synonyms}
\end{table}

\clearpage

\end{document}